\documentclass[10pt,journal,compsoc]{IEEEtran}

\ifCLASSOPTIONcompsoc
  \usepackage[nocompress]{cite}
\else
  \usepackage{cite}
\fi

\usepackage{graphicx}
\usepackage{amsmath,amssymb} % define this before the line numbering.
\usepackage{color}

\usepackage{multirow}

\usepackage{amsmath}

\usepackage{wrapfig}
\usepackage{float}
\usepackage{breqn}

\usepackage{algorithmicx}
\usepackage{algorithm} % http://ctan.org/pkg/algorithms
\usepackage{algpseudocode} %

\usepackage{subfigure}
\usepackage{booktabs}
\usepackage{hyperref}

\usepackage[misc]{ifsym}

\begin{document}

\title{Blind Face Restoration: Benchmark Datasets and a Baseline Model}

% \title{Benchmark Blind Face Restoration and a Baseline Model}

\author{Puyang Zhang,
        Kaihao Zhang,
        Wenhan Luo,
         Changsheng Li,
         and
         Guoren Wang
        \IEEEcompsocitemizethanks{
\IEEEcompsocthanksitem Puyang Zhang, Changsheng Li and Guoren Wang are with the school of computer sciene and technology, Beijing Institute of Technology, Beijing, China. E-mail: \{zhangpuyang2000@foxmail.com;  lcs@bit.edu.cn; wanggrbit@126.com\}  \protect

\IEEEcompsocthanksitem Kaihao Zhang is with the College of Engineering and Computer Science, Australian National University, Canberra, ACT, Australia. E-mail: \{super.khzhang@gmail.com\} \protect

\IEEEcompsocthanksitem Wenhan Luo is with Sun Yat-sen University, China. E-mail: \{whluo.china@gmail.com\}  \protect

\IEEEcompsocthanksitem Corresponding author: Changsheng Li} % <-this % stops a space
\thanks{Manuscript received April 19, 2005; revised August 26, 2015.}}


\markboth{Submitted to IEEE Transactions on Image Processing}%
{Shell \MakeLowercase{\textit{et al.}}: Bare Advanced Demo of IEEEtran.cls for IEEE Computer Society Journals}



\IEEEtitleabstractindextext{%
\begin{abstract}
Blind Face Restoration (BFR) aims to construct a high-quality (HQ) face image from its corresponding low-quality (LQ) input. Recently, many BFR methods have been proposed and they have achieved remarkable success. However, these methods are trained or evaluated on privately synthesized datasets, which makes it infeasible for the subsequent approaches to fairly compare with them. To address this problem, we first synthesize two blind face restoration benchmark datasets called EDFace-Celeb-1M (BFR128) and EDFace-Celeb-150K (BFR512). State-of-the-art methods are benchmarked on them under five settings including blur, noise, low resolution, JPEG compression artifacts, and the combination of them(full degradation). To make the comparison more comprehensive, five widely-used quantitative metrics and two task-driven metrics including Average Face Landmark Distance (AFLD) and Average Face ID Cosine Similarity (AFICS) are applied. Furthermore, we develop an effective baseline model called Swin Transformer U-Net (STUNet). The STUNet with U-net architecture applies an attention mechanism and a shifted windowing scheme to capture long-range pixel interactions and focus more on significant features while still being trained efficiently. Experimental results show that the proposed baseline method performs favourably against the SOTA methods on various BFR tasks. The codes, datasets, and trained models are publicly available at: \url{https://github.com/bitzpy/Blind-Face-Restoration-Benchmark-Datasets-and-a-Baseline-Model}.

\end{abstract}

}

\maketitle

% is not selected <OR> if conference mode is selected - because compsoc
% conference papers position the abstract like regular (non-compsoc)
\IEEEdisplaynontitleabstractindextext

% \fi
\IEEEpeerreviewmaketitle

\section{Introduction}
\label{introduction}
%Among all kinds of images, face images are the most important ones for its unique significance in recording history and maintaining relationships.
Face images are very important in both academic research areas and commercial applications, as they contain rich and unique identity information.
However, in real-world scenarios, face images usually involve complex degradations such as blur, noise, low resolution, compression artifacts, or the combination of them. Face restoration aims at generating a realistic and faithful face image from a degraded one, which can be used for face detection, face recognition, and many other vision tasks. Due to its significant research value and wide range of application scenarios, face restoration has been attracting more and more attention in the image processing and computer vision communities\cite{shen2018deep,chen2019robust,zhang2020occluded}.

For face restoration, traditional methods typically focus on a single type of degradation. Thus face restoration is divided into numerous sub-tasks, including deblurring \cite{shen2018deep}, denoising \cite{yue2019variational,anwar2019real}, super-resolution \cite{dong2015image,chen2018fsrnet,lim2017enhanced,wang2018esrgan,yang2019deep}, compression artifact removal \cite{dong2015compression}. Yet real-world face images are often degraded owing to more than one factor. It is a matter of course that these methods designed for one specific and known degradation perform with limited generalization in the scenery of real-world face images. As such, Blind Face Restoration (BFR) which recovers high-quality (HQ) face images from low-quality (LQ) inputs with unknown degradations is more practical and thus attracts increasing attention in recent years.

\begin{figure*}[t] 
  \centering
  \subfigure[Sample face images from the EDFace-Celeb-1M (BFR128) dataset]{
       \label{fig:BFRBD128} 
      {\includegraphics[width=0.99\linewidth]{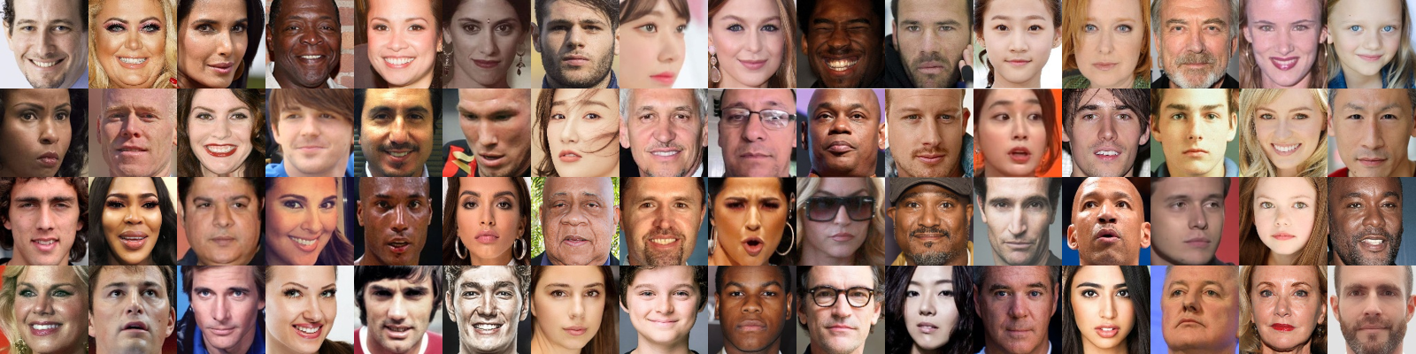}}
  }
  \subfigure[Sample face images from the EDFace-Celeb-150K (BFR512) dataset]{
      \label{fig:BFRBD512} 
      {\includegraphics[width=0.99\linewidth]{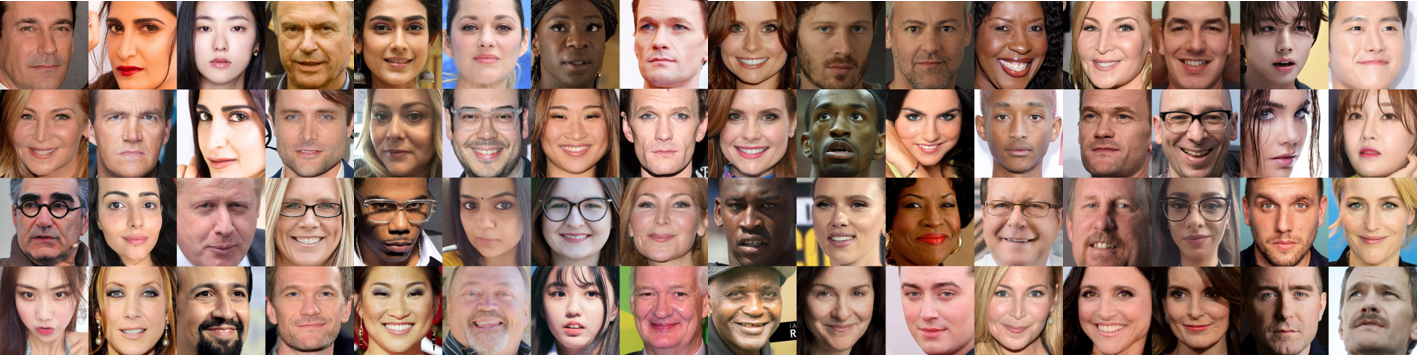}}
  }
  \caption{\textbf{Sample images from the synthesized datasets}. The two datasets contain a large number of face images. We show representative images from subjects of different genders, ages and races.} 
  \label{fig:performance_introduction} 
\end{figure*}

Although the existing BFR methods have demonstrated excellent and remarkable abilities to generate realistic and faithful images in recent years, there are still problems unsolved in the community which hinder the progress of the BFR task. 
First, most of the current BFR methods are trained or evaluated on private facial datasets. This leads to unnecessary and expensive retraining costs.
When a new BFR algorithm is proposed, for a fair comparison, authors have to synthesize datasets by themselves and retrain all previous methods instead of reporting their previous results directly.
Second, the scale of the private datasets used by the existing methods is usually small. This raises an doubt about the generalization of the methods trained with them. 
To advance the development of BFR, we synthesize two benchmark datasets called EDFace-Celeb-1M (BFR128) and EDFace-Celeb-150K (BFR512) based on the EDFace-Celeb dataset \cite{zhang2021edface}. EDFace-Celeb-1M (BFR128) contains about $1.5$ million images of resolution $128\times128$, of which $1.36$ million images are used for training and $145$ thousand images are used for testing. EDFace-Celeb-150K (BFR512) includes about $149$K images with a resolution of $512\times512$. The number of training images is about $132$K and the number of testing is about $17$K.
% To the best of our knowledge, BFR128 and BFR512 are the first large-scale and publicly available datasets for the BFR task. 
Several sample images from them are shown in Fig. \ref{fig:performance_introduction}.    

We present five degradation settings for each of the two datasets to benchmark current representative BFR methods. These settings include blur, noise, low resolution, JPEG compression artifacts, and the combination of them. By evaluating state-of-the-art BFR methods on the two datasets under the above five settings, we gain a more comprehensive understanding of existing BFR methods. Additionally, considering that blind face restoration can be used in many down-stream vision tasks such as face detection and face recognition, we introduce two task-driven metrics, \textit{i.e.}, Average Face Landmark Distance (AFLD) and Average Face ID Cosine Similarity(AFICS), to assess the performance of these methods more objectively and exhaustively.

Finally, we propose a novel Swin Transformer U-Net(STUNet) for the BFR task. Compared with traditional CNN-based models, our STUNet introduces an attention mechanism to enhance the feature representation ability by capturing global interactions between pixels. For most Transformer-based models, the expensive training overhead hinders their usefulness. To this end, we utilize the shifted windowing scheme \cite{liu2021swin} which limits the self-attention computation in a fixed-size local window and meanwhile achieves the cross-window connection. 

% As far as we know, STUNet is the first transformer-based model for BFR.

Overall, the contributions of our work are four-fold:

\begin{itemize}
\item
Firstly, we synthesize two large-scale benchmark datasets for BFR. These two datasets include millions of facial images covering all the typical degradation settings as well as a combination of them. 
%
% To the best of our knowledge, the two datasets proposed in our paper are the first large-scale and public benchmark datasets in the field of BFR. 
On top of our datasets, it is convenient for future research to evaluate the proposed methods and fair to compare with existing methods.

\item

Secondly, we evaluate several representative state-of-the-art BFR methods on two datasets, employing five popularly used quantitative metrics. With such extensive evaluation, we have a clearer understanding of the advantages and disadvantages of the SOTA methods.

\item

Thirdly, we additionally design two task-driven metrics, AFLD and  AFICS, to evaluate the performance of the state-of-the-art BFR methods. This study provides a new perspective to investigate the BFR methods in terms of their influence on down-stream tasks like face detection and recognition.

\item

Fourthly, we propose a baseline model STUNet which adopts the strong Transformer architecture and is capable of enhancing feature representation. Extensive experiments illustrate its superiority on the BFR task. % In addition, this is the first transformer based model for BFR to our knowledge.

\end{itemize}

\section{Related Work}

\subsection{Face Restoration Datasets}

Currently, there are no publicly available datasets specifically for blind face restoration. Most of the face restoration methods use existing face datasets including FFHQ \cite{karras2019style}, CASIA-WebFace \cite{yi2014learning}, VGGFace2 \cite{cao2018vggface2}, IMDB-WIKI \cite{Rothe-ICCVW-2015}, CelebA \cite{liu2015deep}, Helen \cite{le2012interactive}, LWF \cite{huang2008labeled}, BioID \cite{jesorsky2001robust}, AFLW \cite{koestinger2011annotated}, and generate low quality images according to their demands. In 2001, BioID dataset \cite{jesorsky2001robust} is presented and it contains $1,521$ images of $23$ subjects. Huang \emph{et al.} \cite{huang2008labeled} introduce the labeled faces in the wild (LFW) dataset which includes $13,233$ images of $5,749$ subjects. The Annotated Facial Landmarks in the Wild (AFLW) \cite{koestinger2011annotated} dataset is created, including $25,993$ images annotated with up to $21$ landmarks per image. The Helen dataset  consists of $2,330$ high-resolution, accurately labeled face images. The CASIA-WebFace dataset \cite{yi2014learning} is presented in 2014 and it includes $494,414$ images. Liu \emph{et al.} \cite{liu2015deep} introduce the CelebA dataset including $200$K images by labeling images selected from CelebFaces \cite{sun2014deep} and Rothe \emph{et al.} \cite{Rothe-ICCVW-2015} propose the IMDB-WIKI dataset containing $523,051$ face images. More recently, Cao \emph{et al.} \cite{cao2018vggface2} introduce the VGGFace2 dataset which has $3.31$ million images from a large number of subjects. Karras \emph{et al.} \cite{karras2019style} release the FFHQ dataset. It contains $70$K highly varied and high-quality images. See Table \ref{table_dataset} for a clear overview of these datasets.

\begin{table}[!tb]
  \centering
  \caption{\textbf{Representative face datasets used in face restoration tasks}. Specifically for training and evaluating BFR methods, these popularly-used face datasets do not provide public pairs of LQ and HQ images. In the table, "$-$" means that the resolution of the images in the dataset is not fixed.}
  
    \begin{tabular}{ccccc}
    \toprule
    Dataset & Size & Resolution & HQ-LQ \\
    \midrule
    BioID\cite{jesorsky2001robust} & 1521    & $384 \times 288$ & $\times$ \\
    LFW\cite{huang2008labeled}   & 13,233   & $250 \times 250$ & $\times$ \\
    AFLW\cite{koestinger2011annotated}  & 25,993   & $    -         $ & $\times$ \\
    Helen\cite{le2012interactive} & 2330   & $    -          $ & $\times$ \\
    CASIA-WebFace\cite{yi2014learning}  & 494,414   & $250 \times 250    $ & $\times$ \\
    CelebA\cite{liu2015deep} & 200,000  & $  - $ & $\times$ \\
    IMDB-WIKI\cite{Rothe-ICCVW-2015} & 524,230   & $   -       $ & $\times$ \\
    VGGFace2\cite{cao2018vggface2} & 3,310,000 & $      -       $ & $\times$ \\
    FFHQ\cite{karras2019style} & 70,000   &$1024 \times 1024$ & $\times$ \\
    \midrule
    EDFace-Celeb-1M(BFR128) & 1,505,888   &$128 \times 128$ & $\checkmark$ \\
    EDFace-Celeb-150K(BFR512) & 148,962   &$512 \times 512$ & $\checkmark$ \\
    \bottomrule
    \end{tabular}
  \label{table_dataset}
\end{table}

\subsection{Face Restoration Methods}
With the development of deep learning, image restoration has witnessed great success on many tasks, such as image deblurring \cite{zhang2018adversarial,zhang2020deblurring,kupyn2018deblurgan,kupyn2019deblurgan}, image denoising \cite{zhang2017beyond,zhang2018ffdnet,yue2019variational,anwar2019real}, image super-resolution \cite{dong2015image,lim2017enhanced,zhang2018image,ledig2017photo, sajjadi2017enhancenet,wang2018esrgan,zhang2021benchmarking}, image compression artifact removal \cite{fu2019jpeg, dong2015compression}. Face restoration, as an important and popular branch of image restoration, also shows outstanding performance in generating clear and faithful face images in recent years. Thanks to the special facial structure of the face image, there are numerous well-designed algorithms for face restoration. Among these methods, traditional face restoration methods \cite{cao2017attention,chen2018fsrnet,shen2018deep,kim2019progressive,menon2020pulse,huang2017wavelet,zhang2021edface,yu2017hallucinating,yu2018super,dong2015compression,wang2018esrgan} 
pre-define the degradation type before training, thus leading to poor generalization ability. Considering the fact that we cannot know the degradation type when a real-world image is degraded, researchers pay more attention to the so-called blind face restoration (BFR) problem \cite{li2018learning,li2020blind,yang2020hifacegan,yang2021gan,chen2021progressive,wang2021towards}, which is more challenging due to the unknown degradations.

Cao \emph{et al.}\cite{cao2017attention} utilize deep reinforcement learning to sequentially discover attended patches so that high-resolution (HR) faces can be recovered by fully exploiting the global interdependency of the face images. Chen \emph{et al.}\cite{chen2018fsrnet} performer face super-resolution with facial geometry prior. Shen \emph{et al.} \cite{shen2018deep} present a deep convolutional neural network and take the advantage of semantic information to solve the face deblurring problem. Kim \emph{et al.} \cite{kim2019progressive} propose a face super-resolution method that applies a novel facial attention loss to focus on restoring facial attributes. Menon \emph{et al.} \cite{menon2020pulse} propose PULSE, which traverses the high-resolution natural image manifold, searching for images that downscale to the original LR image. When it comes to BFR, Li \emph{et al.} \cite{li2018learning} design a GFRNet including both a wrapping sub-network and a reconstruction sub-network, with a guided image as input for BFR. Moreover, they extend their work by using multi-scale component dictionaries instead of a single reference \cite{li2020blind}. HiFaceGAN, a collaborative suppression and replenishment approach, is introduced by Yang \emph{et al.} \cite{yang2020hifacegan} to progressively replenish facial details. Yang \emph{et al.} \cite{yang2021gan} propose a GAN prior embedded network to generate visually photo-realistic images. Chen \emph{et al.} \cite{chen2021progressive} learn facial semantic prior and make use of multi-scale inputs to recover high-quality images. Wang \emph{et al.} \cite{wang2021towards} leverage pre-trained face GAN to provide diverse and rich priors, solving the contradiction between inaccurate priors offered by low-quality inputs and inaccessible high-quality references.

\subsection{Vision Transformer}
Natural language processing (NLP) model Transformer \cite{NIPS2017_3f5ee243} proposed by Vaswani \emph{et al.} performs favorably against state-of-the-art methods in many NLP tasks. Due to its outstanding ability in feature extraction, Transformer has won great popularity in the computer vision community. A lot of well-designed attention architectures \cite{parmar2018image,hu2019local,ramachandran2019stand,zhao2020exploring,child2019generating,weissenborn2019scaling,ho2019axial,wang2020axial,cordonnier2019relationship} have been proposed for computer vision tasks, advancing the development of vision transformer. However, many of these architectures are not easy to be implemented effectively on hardware accelerators. Dosovitskiy \emph{et al.} \cite{dosovitskiy2020image} introduce Vision Transformer (ViT), which shows more excellent speed-accuracy trade-off on the image classification task than traditional CNN models. To address the limitation that the ViT needs to be trained on large-scale datasets (i.e., JFT-300M) to achieve its best performance, DeiT \emph{et al.} \cite{touvron2021training} propose useful training strategies so that ViT can also be well trained on the smaller ImageNet-1K dataset. Although the result of ViT has been successful, ViT still suffers from the difficulty that the training speed of high-resolution images is unbearably slow for the quadratic increase in complexity with image size. Swin Transformer \cite{liu2021swin} presented by Liu \emph{et al.} whose self-attention is computed with shifted window scheme, greatly improves the training efficiency. Besides, Swin Transformer serving as a general-purpose backbone has become the state-of-the-art method in many vision tasks. Due to the impressive performance of Transformer, there are also several Transformer-based methods \cite{cao2021video,chen2021pre,wang2021uformer,liang2021swinir} for image restoration. Chen \emph{et al.} \cite{chen2021pre} propose an image processing transformer (IPT) which is a pre-trained model and can be employed on intended restoration tasks such as image denoising and image super-resolution after fine-tuning. Cao \emph{et al.} \cite{cao2021video} propose VSR-Transformer that utilizes a self-attention mechanism to restore high-resolution videos from low-resolution inputs. Wang \emph{et al.} \cite{wang2021uformer} present a Uformer using window-based self-attention rather than  global self-attention for image restoration. Liang \emph{et al.} \cite{liang2021swinir} introduce a strong baseline model called SwinIR for image restoration, which is based on the Swin Transformer \cite{liu2021swin}.

\begin{figure}[!tb]
  \centering
  \subfigure[HQ]{
    \label{HQ}
    \includegraphics[width=0.30\linewidth ]{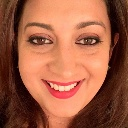}}
    \subfigure[Blur]{
    \label{blur}
    \includegraphics[width=0.30\linewidth ]{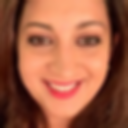}}
    \subfigure[Noise]{
    \label{noise}
    \includegraphics[width=0.30\linewidth ]{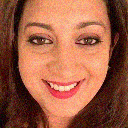}}
    \subfigure[JPEG]{
    \label{jpeg}
    \includegraphics[width=0.30\linewidth ]{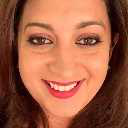}}
    \subfigure[Low Resolution]{
    \label{lr}
    \includegraphics[width=0.30\linewidth ]{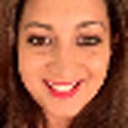}}
    \subfigure[Full]{
    \label{full}
    \includegraphics[width=0.30\linewidth ]{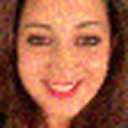}}
\caption{Exemplar images showing the five degradation settings, including blur, noise, JPEG compression, low resolution, and the combination of them (Full), along with the original HQ image.}
  \label{fig:settings} 
\end{figure}

\section{Benchmark Datasets}
\label{sec:benchmark_datasets}

We aim to provide a comprehensive study by quantitatively evaluating the recent state-of-the-art blind face restoration methods. To this end, we need to synthesize appropriate datasets which have not only high-quality images but also their corresponding low-quality images. In this section, we first introduce the construction of the synthesized datasets. And then, we introduce the degradation settings of the two datasets used for both training and evaluation.

\subsection{The Synthesized Datasets}
In both EDFace-Celeb-1M (BFR128) \& EDFace-Celeb-150K (BFR512), we provide high-quality images and for each high-quality image we generate several types of degraded low-quality images. The high-quality images are derived from the EDFace-Celeb \cite{zhang2021edface} dataset. The EDFace-Celeb is a million-scale face image dataset that contains images from more than $20$K different subjects. These subjects are from different races and in different age groups. Each subject has $100\sim1000$ face images from the Internet. 
We select about $1.5$M images whose resolution is $128\times128$ from EDFace-Celeb-1M to synthesize the EDFace-Celeb-1M (BFR128) dataset. Similarly, we construct the EDFace-Celeb-150K (BFR512) dataset by choosing about $150$K images with a resolution is $512\times512$ from EDFace-Celeb-150K. Then we fix the division of training and testing samples in the two datasets. Besides, we synthesize five settings containing pairs of LQ and HQ face images for training and evaluating face restoration methods, which are detailed in the following section. 

\subsection{Image Degradation Settings}
To simulate the complex real-world image low-quality factors, we degrade the quality of HQ images with various types of degradation operations. Following the degradation strategy \cite{yang2020hifacegan}, we employ five degradation settings, called Blur, Noise, JPEG, LR, and Full. Each setting consists of corresponding HQ and LQ image pairs. To obtain the Blur setting, we convolve HQ images with a Gaussian or Motion blur kernel. The images in the Noise setting are generated by adding one of Gaussian, Laplace, and Poisson noise to the HQ images. Compressing HQ images by JPEG, the JPEG setting is created. For the LR setting, we use Bicubic interpolation with a random downsampling factor between 2 and 8 to synthesize low-resolution images. Besides, the Full setting images are synthesized by randomly superimposing the above four types of degradations over HQ images. These settings correspond to the face deblurring, denoising, JPEG compression artifact removal, super-resolution, and blind restoration tasks. Our synthetic datasets are sufficiently challenging so as to fulfill the goal of training BFR approaches in the real-world scenery.
Fig. \ref{fig:settings} shows exemplar images of an original HQ image and its LQ counterparts by different degradation operations.

\section{STUNet}
\begin{figure*}[!tb]
  \centering
      {\includegraphics[width=0.99\linewidth]{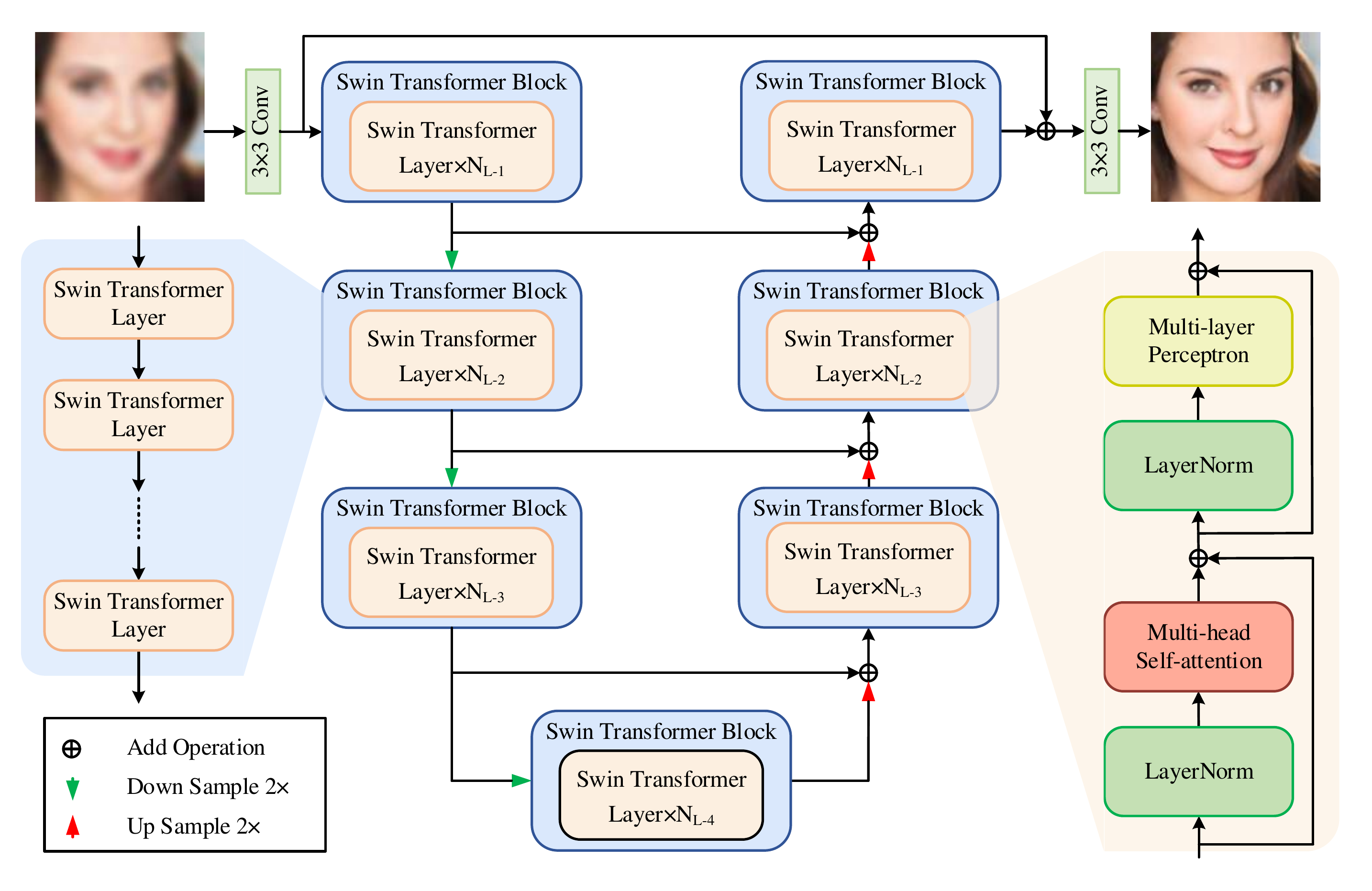}}
\caption{\textbf{The architecture of the proposed STUNet for blind face restoration}. STUNet takes a low-quality image as input, then uses a convolutional layer to extract shallow features. Next, we use several Swin Transformer blocks to extract deep features further. Finally, a convolutional layer is used to recover a high-quality face image from the deep features.}
\label{fig:network} 
\end{figure*}

Blind face restoration is a challenging task because of its complex degradation of input images. Existing BFR methods are typically based on a convolutional neural network. While currently Swin Transformer \cite{liu2021swin} based methods have been achieving state-of-the-art performance in many computer vision tasks. We thus propose a novel baseline model for blind face restoration, which uses Swin Transformer as the backbone instead of a convolutional neural network. We aim to design a lightweight but promising network architecture that can be conveniently used and benefit future research. In the following, we introduce our proposed baseline model Swin Transformer U-net (STUNet).

\subsection{Network Architecture}
As shown in Fig. \ref{fig:network}, the proposed STUNet consists of three parts: preprocessing embedding module, feature extraction module, and image restoring module.

\textbf{Preprocessing Embedding Module}. Given a low-quality input image $I_{LQ}{\in} \mathbb{R}^{H{\times}W{\times}C_I}$ ($H$ is the image height, $W$ is the image width, and $C_I$ is the image channel number), our STUNet first applies a $3\times3$ convolutional layer to extract a shallow feature embedding as
\begin{equation}
F_{C0}=H_{C0}(I_{LQ}),
\end{equation}
where $F_{C0}{\in}\mathbb{R}^{H{\times}W{\times}C}$ is the features extracted via the $3\times3$ convolutional layer, $C$ is the channel number of the feature, and $H_{C0}$ denotes the function representing the first convolutional layer. The convolutional layer works well in mapping low dimensional image space to high dimensional feature space in a simple yet effective way.

\textbf{Feature Extraction Module}. The low-level features from the preprocessing embedding module are then processed by a feature extraction module to obtain deep features.
The feature extraction module is a symmetric 4-level encoder-decoder architecture composed of Swin Transformer blocks (STB) at different levels. In the encoder part, the shallow feature $F_{C0}$ is first fed to a Swin Transformer block to extract features $F_{e1}{\in}\mathbb{R}^{H{\times}W{\times}C}$ by
\begin{equation}
F_{e1}=H_{e1}(F_{C0}),
\end{equation}
where $H_{e1}$ is the function to represent the process by the Swin Transformer block at the first level. Next, we apply the pixel-unshuffle operation to downsample the feature $F_{e1}$, and then we obtain deeper features  $F_{e2}{\in}\mathbb{R}^{\frac{H}{2}{\times}\frac{W}{2}{\times}2C}$ via a Swin Transformer block as
\begin{equation}
F_{e2}=H_{e2}(H_{down}(F_{e1})),
\end{equation} 
where $H_{down}$ and $H_{e2}$ denote the downsampling function and the process by the Swin Transformer block at the second level.

Similarly, the encoder further extracts the latent features $F_{e3}{\in}\mathbb{R}^{\frac{H}{4}{\times}\frac{W}{4}{\times}4C}$ and $F_{e4}{\in}\mathbb{R}^{\frac{H}{8}{\times}\frac{W}{8}{\times}8C}$ as the following, \begin{equation}
F_{e3}=H_{e3}(H_{down}(F_{e2})),
\end{equation} 

\begin{equation}
F_{e4}=H_{e4}(H_{down}(F_{e3})),
\end{equation} 
where $H_{e3}$, $H_{e4}$ are the functions representing Swin Transformer block at level-3 and level-4, respectively. 

From the top to bottom level, the encoder converts the high-resolution shallow features with a small number of channels to the low-resolution latent feature with a large number of channels. Next, the decoder progressively reconstructs the high-resolution features from the low-resolution input $F_{e4}$ in a symmetrical way with the encoder. Besides, in order to extract more powerful features, we apply skip-connections to aggregate the encoder and decoder features. In addition, we use the pixel-shuffle operation for upsampling. Equations \ref{equ:decode1},  \ref{equ:decode2}, \ref{equ:decode3} illustrate the whole decoding process as 
\begin{equation}
F_{d3}=H_{d3}(H_{up}(F_{e4})+F_{e3}),
\label{equ:decode1}
\end{equation}

\begin{equation}
F_{d2}=H_{d2}(H_{up}(F_{d3})+F_{e2}),
\label{equ:decode2}
\end{equation} 

\begin{equation}
F_{d1}=H_{d1}(H_{up}(F_{d2})+F_{e1}).
\label{equ:decode3}
\end{equation} 
In these equations, $F_{d3}{\in}\mathbb{R}^{\frac{H}{4}{\times}\frac{W}{4}{\times}4C}$, $F_{d2}{\in}\mathbb{R}^{\frac{H}{2}{\times}\frac{W}{2}{\times}2C}$, and $F_{d1}{\in}\mathbb{R}^{H{\times}W{\times}C}$ are the latent features with different levels of receptive field. $H_{d3}$, $H_{d2}$, and $H_{d1}$ denote the functions representing the level-3, level-2, and level-1 Swin Transformer blocks, respectively. $H_{up}$ denotes the upsampling function.

\textbf{Image Restoration Module}. Finally, the network adds the shallow features to deep features and takes the aggregated features as input to restore the high-quality image via a $3\times3$ convolutional layer. It is formulated by the following two equations.
\begin{equation}
F_{agg} = F_{C0}+F_{d1},
\end{equation} 
\begin{equation}
I_{HQ}= H_{res}(F_{agg}),
\end{equation} 
where $F_{agg}$, $I_{HQ}{\in}\mathbb{R}^{H{\times}W{\times}CI}$ denote the aggregated features and the high-quality image respectively. $H_{res}$ is the function representing the last convolutional layer
The final output of STUNet is the HQ image $I_{HQ}$.

\subsection{Swin Transformer Block}
Our Swin Transformer Block (STB) consists of several Swin Transformer layers. Given the input feature $F_{li,0}$ of the Swin Transformer block at level-i, the Swin Transformer block applies $N_{li}$ Swin Transformer layers to extract deep features $F_{li,1}$, $F_{li,2}$, $F_{li,3}$,...,$F_{li,N_{li}}$ as

\begin{equation} 
F_{li,j}=H_{STL_{li},j}(F_{li,j-1}), \ i=1,2,3,4, \ j=1,2,...,N_{li}, 
\end{equation}
where $H_{STL_{li},j}$ denotes the function indicating the j-th Swin Transformer layer in the Swin Transformer block at level-i and $F_{li,N_{li}}$ is the output feature of this Swin Transformer block. From the top to the bottom level, we gradually increase the number of Swin Transformer layers in the Swin Transformer block. The appropriate number of Swin Transformer layers at different levels ensures the efficiency of the network. Table \ref{table:STB} shows the extracted feature size and the number of Swin Transformer layers in the Swin Transformer blocks at different levels in detail. 
%Overall, our design for the Swin Transformer block mainly has two advantages. 
%\begin{itemize}
%\item
%First, we improve the feature representation ability by applying the self-attention mechanism.  
%\item
%Second, the appropriate number of Swin Transformer Layer at different levels and shifted windowing computing scheme ensure the efficiency of the network.
%\end{itemize}

%\fixme{these so-called advantages are not even mentioned in the STB. For example, the self attention in point 1, and the shifted windowing scheme in point 2. I notice you provide these points in section 4.3. So I would suggest you move the claims of these two advantages to the end of section 4.3.}
%{\color{red} However, the Swin Transformer layer is proposed by \cite{liu2021swin}, it is not proposed by us}

\begin{table}[!tb]
  \centering
  \caption{The extracted feature size and the number of Swin Transformer layers in Swin Transformer blocks at different levels.}
    \begin{tabular}{cccccc}
    \toprule
    STB Level & Feature Size & STL Number\\
    \midrule
    Level-1 & $H{\times}W{\times}C $ & 4\\
    Level-2 & $H/2{\times}W/2{\times}2C$ & 6\\
    Level-3 & $H/4{\times}W/4{\times}4C$ & 6\\
    Level-4 & $H/8{\times}W/8{\times}8C$ & 8\\
    
    \bottomrule
    \end{tabular}
  \label{table:STB}
\end{table}

\noindent{\textbf{Swin Transformer Layer (STL)} \cite{liu2021swin} is a variant of the original Transformer layer \cite{vaswani2017attention}. It applies multi-head self-attention in a fixed-size local window rather than the global feature. 
Given an input feature $F_0{\in}\mathbb{R}^{H{\times}W{\times}C}$, the Swin Transformer layer divides it into local windows and converts the feature dimension from 3D to 2D, so that it obtains $\frac{HW}{N^2}$ local features whose size is ${N^2}{\times}C$. 
For each local feature, it operates the standard self-attention mechanism as
\begin{equation} 
Attention(Q,K,V)=Softmax(\frac{QK^T}{\sqrt{d}}+B)V, 
\end{equation} 
where $Q, K, V{\in}\mathbb{R}^{N^2{\times}d}$ denote the query, key, and value matrices; $B$ denotes the learnable parameters corresponding to the relative position. Then it implements the multi-head self-attention (MSA)
\cite{vaswani2017attention} by performing the attention function in parallel and concatenating the output. Besides, because it limits the multi-head self-attention to the fixed-size local window, there is no connection among features in different local windows. To overcome this limitation, it applies the shifted window mechanism to realize the cross-window connection. In addition, the Swin Transformer layer applies a multi-layer perceptron (MLP) which is composed of two fully-connected layers with GELU non-linearity function in between to further extract features. Before every  MSA and MLP module, it sets a LayerNorm(LN) module to ensure the stability of feature distribution, and after each of them, it applies the residual connection. Equation \ref{equ:MSA} and \ref{equ:MLP} illustrate the complete process of the Swin Transformer layer as

\begin{equation}
F_{MSA}=H_{MSA}(H_{LN}(F_{local}))+F_{local}, 
\label{equ:MSA}
\end{equation} 

\begin{equation}
F_{MLP}=H_{MLP}(H_{LN}(F_{MSA}))+F_{MSA},
\label{equ:MLP}
\end{equation}
where $F_{local}$, $F_{MSA}$ and $F_{MLP}$ are the local feature input, the output of the MSA module, and the output of MLP module; $H_{LN}$, $H_{MSA}$, $H_{MLP}$ are the functions indicating LayerNorm,  multi-head self-attention, and multi-layer perceptron. $F_{MLP}$ is the final output of the Swin Transformer layer.}

\subsection{Loss Function and Optimization}
Our STUNet applies the $L1$ pixel loss as the loss function,
\begin{equation} 
\mathcal{L}={||I_{HQ}-I_{GT}||}_1
\end{equation}
where the $I_{HQ}$ is the high-quality output image of STUNet taking the $I_{LQ}$ as input and the $I_{GT}$ is the ground-truth image corresponding to $I_{LQ}$. We optimize the parameters of the network by minimizing the loss $\mathcal{L}$ via stochastic gradient descent. The whole training process of the proposed STUNet algorithm is shown in Algorithm \ref{algorithm:train}.

\begin{algorithm}[t]
	\caption{Training Process of STUNet} 
	\textbf{Input:} the low-quality image set $\mathcal{x}$, the ground-truth image set $\mathcal{Y}$ and the epoch number;\\
    \textbf{Output:} the trained parameters of model $\Psi'$;
	\begin{algorithmic}[1]
	\State{Initialize $\Psi'$;}
	\For{each epoch}
    	\For{each iteration}
    	    \State{generate the output of the model};
    	    \State{calculate the $L1$ loss $\mathcal{L}$ };
    	    \State{update the parameter of $\Psi'$ by minimizing $\mathcal{L}$ };
    	\EndFor
    \EndFor
    \end{algorithmic} 
\label{algorithm:train} 
\end{algorithm} 

\section{Experiments and Analysis}

\begin{table*}
\centering
\caption{\textbf{Performance comparison of representative BFR methods on the from EDFace-Celeb-1M (BFR128) dataset}. Results are reported in terms of PSNR, SSIM, MS-SSIM, LPIPS, and NIQE metrics. The best and the second best performance results are marked with red and blue colors, respectively.}
\begin{tabular}{c|c|ccccc} 
\hline
Task                   & Methods & PSNR↑ & SSIM↑ & MS-SSIM↑ & LPIPS↓  & NIQE↓  \\ 
\hline
\multirow{6}{*}{Face Deblurring}  
                       & HiFaceGAN \cite{yang2020hifacegan} &22.4598&0.7974 & 0.9420         &0.0739 & {\color{blue}8.7261}\\
                       & PSFR-GAN \cite{chen2021progressive}  &{\color{red}29.1411} &{\color{red}0.8563}& {\color{red}0.9818}       & {\color{blue}0.0480} &9.0008      \\
                       & GFP-GAN \cite{wang2021towards}    &25.3822 &0.7461&0.9534    & {\color{red}0.0704}    &12.3608  \\
                       & GPEN \cite{yang2021gan}   & 24.9091 &0.7307 &  0.9500     &0.0887  &{\color{red}8.2288} \\
                       & STUNet   &{\color{blue}27.3912} & {\color{blue}0.8080} & {\color{blue}0.9669}     & 0.2019   &  12.2652    \\ 
\hline
\multirow{6}{*}{Face Denoising} 
                        & HiFaceGAN \cite{yang2020hifacegan} &26.2976&0.8801 &0.9663       &0.0306 & {\color{red}7.2432}\\
                       & PSFR-GAN \cite{chen2021progressive} &{\color{blue}33.1007} &0.8563& 0.9818      & 0.0480 &9.0008      \\
                       & GFP-GAN \cite{wang2021towards}   &31.1053 &0.8802&0.9849 & {\color{blue}0.0234}    &{\color{blue}7.9522}  \\
                       & GPEN \cite{yang2021gan}  &33.0744 &{\color{blue}0.9086}&{\color{blue}0.9871}     & {\color{red}0.0211}    &8.0616  \\
                       & STUNet   & {\color{red}34.8914} &{\color{red}0.9302} & {\color{red}0.9900 }    &    0.0331 & 8.5349      \\ 
\hline
\multirow{6}{*}{Face Artifact Removal} 
                        & HiFaceGAN \cite{yang2020hifacegan} &23.8228&0.8531 & 0.9567    &0.0453 & {\color{red}7.6479}\\
                       & PSFR-GAN \cite{chen2021progressive} &{\color{blue}31.9455} &{\color{blue}0.8899}& {\color{blue}0.9887}   & {\color{red}0.0190} &8.3158     \\
                       & GFP-GAN \cite{wang2021towards}   &31.0910 &0.8804&0.9874& {\color{blue}0.0227}   &{\color{blue}7.8027 } \\
                       & GPEN \cite{yang2021gan}  & 30.5753 &0.8556 &0.9837         &0.0241  &7.8074 \\
                       & STUNet   &{\color{red}33.2082} & {\color{red}0.9171} & {\color{red}0.9912}     &  0.0582   & 10.5596      \\ 
\hline
\multirow{6}{*}{Face Super Resolution}   
                        & HiFaceGAN \cite{yang2020hifacegan}&24.2965&{\color{blue}0.7792} & {\color{blue}0.9493}         &0.0911 & 8.4801\\
                       & PSFR-GAN \cite{chen2021progressive} &23.9671 &0.6858& 0.9381      & 0.1364 &{\color{red}7.4807}      \\
                       & GFP-GAN \cite{wang2021towards}   &{\color{blue}25.7118} &0.7558&0.9492     & {\color{red}   0.0762} &11.4428  \\
                       & GPEN \cite{yang2021gan}   &25.0208 &0.7306&0.9448     & {\color{blue}0.0843}    &{\color{blue}7.9052}  \\
                       & STUNet   &{\color{red}27.1206} & {\color{red}0.8037} & {\color{red}0.9566}     &   0.2018  & 12.7177 \\ 
\hline
\multirow{6}{*}{Blind Face Restoration}  
                        & HiFaceGAN \cite{yang2020hifacegan}&22.2179&{\color{red}0.7088} & 0.9128         &0.1528 & 9.6864\\
                       & PSFR-GAN \cite{chen2021progressive} &22.2620 &0.5199& 0.8811      & 0.3558 &{\color{blue}8.3706}      \\
                       & GFP-GAN \cite{wang2021towards}  &{\color{blue}23.4159} &0.6707&{\color{blue}0.9185}  &{\color{red}0.1354} &12.6364  \\
                       & GPEN \cite{yang2021gan}  & 22.9731 &0.6348 &0.9119         &{\color{blue}0.1387}   &{\color{red}8.0709} \\
                       & STUNet   &{\color{red}24.5500} & {\color{blue}0.6978} & {\color{red}0.9225}&  0.3523   & 13.0601    \\ 
\hline
\end{tabular}
\label{table:128_1}
\end{table*}

In this section, we benchmark the existing BFR methods along with our proposed STUNet, on both datasets. All the benchmark studies are conducted under the five challenging settings described in Sec. \ref{sec:benchmark_datasets}.

\subsection{Evaluated BFR Methods}
In this benchmark study, we evaluate the performance of five state-of-the-art BFR methods including HiFaceGAN \cite{yang2020hifacegan}, PSFR-GAN \cite{chen2021progressive}, GFP-GAN \cite{wang2021towards}, GPEN \cite{yang2021gan}, and DFDNet \cite{li2020blind}. 

\subsection{Implementation}
Both the synthesized datasets are associated with five degradation settings corresponding to five face restoration tasks. Each setting corresponds to pairs of low-quality and high-quality images, which are the input and ground truth images. 

Among the methods in our benchmark study, the DFDNet does not release the training code, so we use the pre-trained model provided by the authors. 
Because this pre-trained model requires the input image with a resolution of $512{\times}512$, we only evaluate its performance on the EDFace-Celeb-150K (BFR512) dataset.
For the other methods, we use the released source code from the original publication to train and evaluate them on the synthesized datasets. To ensure the fairness of the comparison, we fix the number of training epochs and learning rate for all methods, which are set as $3$ and $0.001$, respectively. All the models are trained using 3090 GPU. 
To benchmark these methods more comprehensively and convincingly, we use both full-reference and non-reference quantitative metrics. The full-reference quantitative metrics we employed include PSNR, SSIM, and MS-SSIM. 
The non-reference quantitative metrics include NIQE and LPIPS. 
In addition, considering the popular application of face restoration in other vision tasks, we also propose two task-driven metrics: average face landmark distance (AFLD) and average face ID cosine similarity (AFICS) to further assess the performance of these methods. Note that all the metrics are calculated in the RGB space.

\begin{table}
\centering
\caption{\textbf{Performance comparison of representative BFR methods on the from EDFace-Celeb-1M (BFR128) dataset}. Results are reported in terms of two task-driven metrics including AFLD and AFICS. 
The best and the second best performance results are marked with red and blue colors, respectively.}
\begin{tabular}{c|c|cc} 
\hline
Task                   & Methods & AFLD↓ & AFICS↑  \\ 
\hline
\multirow{2}{*}{Face Deblurring}  
                       & HiFaceGAN \cite{yang2020hifacegan} &{\color{blue}0.0286}&{\color{blue}0.8981} \\
                       & PSFR-GAN \cite{chen2021progressive}  & {\color{red}0.0236}&{\color{red}0.9402} \\
                       & GFP-GAN \cite{wang2021towards}    &0.0305      &0.8672          \\
                       & GPEN \cite{yang2021gan}   & 0.0304     &0.8357            \\
                       & STUNet   &0.0288      &0.8609            \\ 
\hline
\multirow{2}{*}{Face Denoising} 
                       & HiFaceGAN \cite{yang2020hifacegan}& 0.0237 &0.9297           \\
                       & PSFR-GAN \cite{chen2021progressive} & 0.0222 &{\color{blue}0.9541}           \\
                       & GFP-GAN \cite{wang2021towards}   &0.0223      &0.9510          \\
                       & GPEN \cite{yang2021gan}  & {\color{blue}0.0221}&0.9530            \\
                       & STUNet   &{\color{red}0.0211} &{\color{red}0.9619}\\ 
\hline
\multirow{2}{*}{Face Artifact Removal} 
                       & HiFaceGAN \cite{yang2020hifacegan}& 0.0252& 0.9459          \\
                       & PSFR-GAN \cite{chen2021progressive} &{\color{blue}0.0215}&{\color{blue} 0.9788}  \\
                       & GFP-GAN \cite{wang2021towards}   &0.0217      &0.9778          \\
                       & GPEN \cite{yang2021gan}  &0.0222      &0.9696            \\
                       & STUNet   &{\color{red}0.0206} &{\color{red}0.9804}        \\ 
\hline
\multirow{2}{*}{Face Super Resolution}   
                       & HiFaceGAN \cite{yang2020hifacegan}&{\color{red}0.0268}&{\color{red}0.8129}\\
                       & PSFR-GAN \cite{chen2021progressive} &0.0328      &0.7294           \\
                       & GFP-GAN \cite{wang2021towards}   & {\color{blue}0.0288}&0.7759\\
                       & GPEN \cite{yang2021gan}  &0.0294 &0.7583            \\
                       & STUNet   &  0.0297    &{\color{blue}0.7859}            \\ 
\hline
\multirow{2}{*}{Blind Face Restoration}  
                        & HiFaceGAN \cite{yang2020hifacegan} &{\color{red}0.0342}&{\color{blue}0.6137}\\
                       & PSFR-GAN \cite{chen2021progressive} & 0.0445     &0.5519           \\
                       & GFP-GAN \cite{wang2021towards}   &0.0358      &0.5931          \\
                       & GPEN \cite{yang2021gan}  &{\color{blue}0.0357}     &0.5769    \\
                       & STUNet   &0.0406      &{\color{red}0.6264}            \\ 
\hline

\end{tabular}

\label{table:128_2}
\end{table}

\subsection{EDFace-Celeb-1M (BFR128) Benchmark Results}

\begin{figure*}[!tb]
  \centering
    \subfigure[Visual results of different methods at \textbf{Face Deblurring} task on the EDFace-Celeb-1M (BFR128) dataset]{
    \label{128_blue}
    \includegraphics[width=0.99\linewidth ]{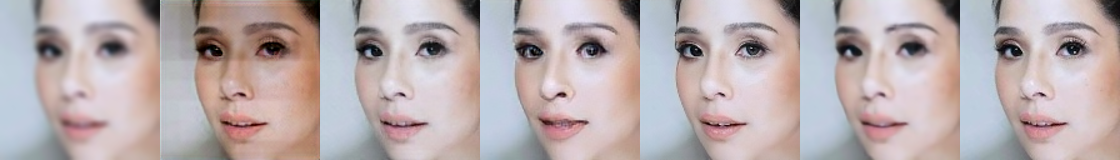}}
    \subfigure[Visual results of different methods for the \textbf{Face Denoising} task on the EDFace-Celeb-1M (BFR128) dataset]{
    \label{128_noise}
    \includegraphics[width=0.99\linewidth ]{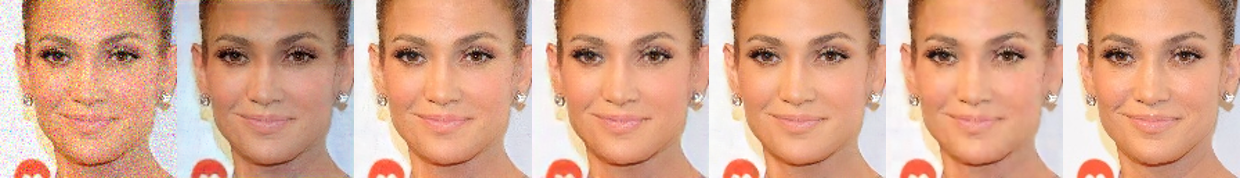}}
    \subfigure[Visual results of different methods for the \textbf{Face Artifact Removal} task on the EDFace-Celeb-1M (BFR128) dataset]{
    \label{128_jpeg}
    \includegraphics[width=0.99\linewidth ]{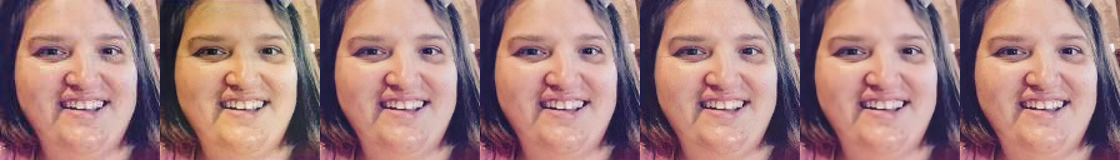}}
    \subfigure[Visual results of different methods for the \textbf{Face Super Resolution} task on the EDFace-Celeb-1M (BFR128) dataset]{
    \label{128_lr}
    \includegraphics[width=0.99\linewidth ]{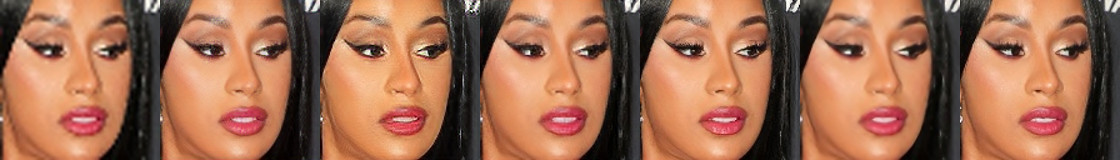}}
    \subfigure[Visual results of different methods for the \textbf{Blind Face Restoration} task on the EDFace-Celeb-1M (BFR128) dataset]{
    \label{128_full}
   \includegraphics[width=0.99\linewidth ]{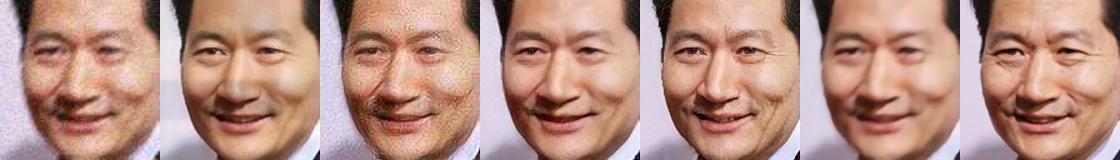}}
\caption{ \textbf{Visual results corresponding to different tasks on the EDFace-Celeb-1M (BFR128) dataset. } From left to right: LQ image, results of HiFaceGAN\cite{yang2020hifacegan}, PSFR-GAN\cite{chen2021progressive}, GFP-GAN\cite{wang2021towards}, GPEN\cite{yang2021gan}, STUNet and HQ image.}
  \label{fig:128_results} 
\end{figure*}

\begin{table*}
\centering
\caption{\textbf{Performance comparison of representative BFR methods on the EDFace-Celeb-150K (BFR512) dataset}. Results are reported in terms of PSNR, SSIM, MS-SSIM, LPIPS, and NIQE metrics. The best and the second best performance results are marked with red and blue colors, respectively.}
\begin{tabular}{c|c|ccccc} 
\hline
Task                   & Methods & PSNR↑ & SSIM↑ & MS-SSIM↑ & LPIPS↓  & NIQE↓  \\ 
\hline
\multirow{6}{*}{Face Deblurring}  & DFDNet \cite{li2020blind}  &25.4072&0.6512&0.8724 &0.4008 & {\color{red}7.8913}\\
                       & HiFaceGAN \cite{yang2020hifacegan} &26.7421 &{\color{red} 0.8095}&{\color{red}0.9382} &{\color{blue}0.2029 }&16.6642 \\
                       & PSFR-GAN \cite{chen2021progressive}  &27.4023  &0.7604 & 0.9155 &0.2292  &17.4076      \\
                       & GFP-GAN \cite{wang2021towards}    &{\color{blue}28.8166} & 0.7709&0.9180 &{\color{red}0.1721}&15.5942        \\
                       & GPEN \cite{yang2021gan}   &27.0658&0.7175 &0.8928&0.2188 &15.3187       \\
                       & STUNet &{\color{red}29.5572}&{\color{blue}0.8052} &{\color{blue}0.9289} &0.3381&{\color{blue}14.7874}       \\ 
\hline
\multirow{6}{*}{Face Denoising} & DFDNet \cite{li2020blind}&24.3618&0.5738 &0.8423&0.3238&{\color{red}7.7809}\\
                       & HiFaceGAN \cite{yang2020hifacegan} &30.0409&{\color{blue}0.8731}&0.9563 &{\color{blue}0.1439} &16.7363\\
                       & PSFR-GAN \cite{chen2021progressive} &28.5397&0.8232 &0.9390 &0.2208& 19.4719      \\
                       & GFP-GAN \cite{wang2021towards} &{\color{blue}33.2020} &0.8711&{\color{blue}0.9582} &{\color{red}0.1259} &15.8440\\
                       & GPEN \cite{yang2021gan} &32.3736&0.8517 &0.9506 &0.1555 &{\color{blue}15.6820} \\
                       & STUNet &{\color{red}34.5500} &{\color{red}0.8848}&{\color{red}0.9587} &0.1787 & 16.5480      \\ 
\hline
\multirow{6}{*}{Face Artifact Removal}&DFDNet \cite{li2020blind}&27.4781&0.7845&0.9409 &0.2241&{\color{red}7.5553} \\
                       & HiFaceGAN \cite{yang2020hifacegan} &27.1164 & 0.8897&0.9635 &0.1241&18.7117  \\
                       & PSFR-GAN \cite{chen2021progressive} &29.4285 &0.9101 &0.9719 &0.1245 &{\color{blue}15.9760} \\
                       & GFP-GAN \cite{wang2021towards}   &{\color{blue}35.7201} &{\color{blue}0.9144} & {\color{blue}0.9780}&{\color{red} 0.0842} &16.8320   \\
                       & GPEN \cite{yang2021gan}  &33.8355 &0.8701 &0.9657  &{\color{blue}0.0986} &16.9854  \\
                       & STUNet   &{\color{red}36.5017} &{\color{red} 0.9246}& {\color{red}0.9799}&0.1411&16.0487       \\ 
\hline
\multirow{6}{*}{Face Super Resolution}& DFDNet \cite{li2020blind}&26.8691&0.7405&0.9224& 0.2620& {\color{red}7.4796}\\
                       & HiFaceGAN \cite{yang2020hifacegan} &26.6103 & 0.8480&0.9476 &0.1681 &15.8911 \\
                       & PSFR-GAN \cite{chen2021progressive} &33.1233  &0.8588 &0.9602  &{\color{blue}0.1331} &16.7143\\
                       & GFP-GAN \cite{wang2021towards}   &{\color{blue}33.4217}  &{\color{blue}0.8629}&{\color{blue}0.9610} &{\color{red}0.1127} &16.8970       \\
                       & GPEN \cite{yang2021gan}  &31.3507&0.8273 &0.9501 &0.1357&{\color{blue}15.7813}  \\
                       & STUNet   &{\color{red}33.9060} &{\color{red}0.8809} &{\color{red}0.9636} &0.2235 &17.0899       \\ 
\hline
\multirow{6}{*}{Blind Face Restoration} & DFDNet \cite{li2020blind}&23.9349&0.5573&0.8053& 0.4231&{\color{red}9.0084}\\
                       & HiFaceGAN \cite{yang2020hifacegan} &25.3083 & 0.7260&{\color{blue}0.8701} &{\color{blue}0.3012} &14.7883 \\
                       & PSFR-GAN \cite{chen2021progressive} &26.2998 &0.6934 & 0.8581&0.3167 &17.1906  \\
                       & GFP-GAN \cite{wang2021towards}  &{\color{red}28.4809}  & {\color{red}0.7857}&{\color{red}0.9255}  &{\color{red}0.2171} &{\color{blue}14.4933}  \\
                       & GPEN \cite{yang2021gan}  &25.5778 &0.6721 &0.8448 &0.3113 & 15.8422      \\
                       & STUNet   &{\color{blue}27.1833} &{\color{blue}0.7346} &0.8654 &0.4457& 17.0305 \\ 
\hline

\end{tabular}
\label{table:512_1}
\end{table*}

\begin{table}[t]
\centering
\caption{\textbf{Performance comparison of representative BFR methods on the EDFace-Celeb-150K (BFR512) dataset}. Results are reported in terms of two task-driven metrics including AFLD and AFICS. The best and the second best performance results are marked with red and blue colors, respectively.}

\begin{tabular}{c|c|cc} 
\hline
Task                   & Methods &AFLD↓ & AFICS↑   \\ 
\hline
\multirow{6}{*}{Face Deblurring}  & DFDNet \cite{li2020blind}    &0.0221 &0.8618            \\
                       & HiFaceGAN \cite{yang2020hifacegan} &{\color{red}0.0177} &{\color{blue}0.9673}         \\
                       & PSFR-GAN \cite{chen2021progressive}  &0.0191 & 0.9609          \\
                       & GFP-GAN \cite{wang2021towards}    &{\color{blue}0.0186}& {\color{red}0.9714}          \\
                       & GPEN \cite{yang2021gan}   &0.0201& 0.9554           \\
                       & STUNet   &0.0207  &0.9644            \\ 
\hline
\multirow{6}{*}{Face Denoising} & DFDNet \cite{li2020blind}   &0.0262&0.9062            \\
                       & HiFaceGAN \cite{yang2020hifacegan}&{\color{red}0.0168} &0.9650  \\
                       & PSFR-GAN \cite{wang2021towards}  &0.0175 &0.9575       \\
                       & GFP-GAN \cite{wang2021towards}  &{\color{blue}0.0169} &{\color{blue}0.9657}          \\
                       & GPEN \cite{yang2021gan}  &0.0173& 0.9623           \\
                       & STUNet   &0.0450  &{\color{red}0.9708}  \\ 
\hline
\multirow{6}{*}{Face Artifact Removal}  & DFDNet \cite{li2020blind}   &0.0199&0.9433            \\
                       & HiFaceGAN \cite{yang2020hifacegan} &0.0166  & 0.9743          \\
                       & PSFR-GAN \cite{wang2021towards}  &{\color{red}0.0161} &0.9748           \\
                       & GFP-GAN \cite{wang2021towards}   &{\color{blue}0.0162} &{\color{blue}0.9792}          \\
                       & GPEN \cite{yang2021gan}  &0.0167& 0.9751           \\
                       & STUNet   &0.0175 &{\color{red}0.9794}            \\ 
\hline
\multirow{6}{*}{Face Super Resolution}     & DFDNet \cite{li2020blind}   &0.0201 &0.9431            \\
                       & HiFaceGAN \cite{yang2020hifacegan} &0.0173  & 0.9584          \\
                       & PSFR-GAN \cite{wang2021towards}  &{\color{blue}0.0166} & 0.9721          \\
                       & GFP-GAN \cite{wang2021towards}   &{\color{red}0.0164} &{\color{blue}0.9745}         \\
                       & GPEN \cite{yang2021gan}  & 0.0172& 0.9698           \\
                       & STUNet   &0.0174 &{\color{red}0.9763}            \\ 
\hline
\multirow{6}{*}{Blind Face Restoration}  & DFDNet \cite{li2020blind}   &0.0247&0.8440            \\
                       & HiFaceGAN \cite{yang2020hifacegan} &{\color{blue}0.0208} &0.8918          \\
                       & PSFR-GAN \cite{wang2021towards}  &0.0212 &0.8964           \\
                       & GFP-GAN \cite{wang2021towards}   &{\color{red}0.0186} &{\color{red}0.9536}         \\
                       & GPEN \cite{yang2021gan}  &0.0221 &0.8782            \\
                       & STUNet   &0.0326 &{\color{blue}0.9045}            \\ 
\hline

\end{tabular}
\label{table:512_2}
\end{table}

We first evaluate these methods and our proposed STUNet on the EDFace-Celeb-1M (BFR128) to study their performance in the scenarios of several types of degradations when the low-quality input images are of resolution $128 \times 128$.

Table \ref{table:128_1} reports the comparison results of five methods, including our proposed STUNet.
In the table, Face Deblurring, Face Denoising, Face Artifact Removal, Face Super-Resolution, and Blind Face Restoration are the five face restoration tasks corresponding to the degradation settings of Blur, Noise, JPEG, LR, and Full, respectively. 
PSNR, SSIM, and MS-SSIM are the three full-reference quantitative metrics.
LPIPS and NIQE are the two non-reference quantitative metrics. 
The results in red color and blue color represent the best and the second best results. 

We have the following findings regarding the results of different methods on the five tasks from Table \ref{table:128_1}.
1) In terms of the full-reference quantitative metrics including PSNR, SSIM, and MS-SSIM, our STUNet achieves the best (face denoising, face artifact removal, and face super resolution) or the second best performance (face deblurring and blind face restoration), consistently beating other compared methods. Additionally, PSFR-GAN performs best for the task of face deblurring, and the second best for the face artifact removal task. HiFaceGAN, GFP-GAN and GPEN perform favourably, but not as well as STUNet and PSFR-GAN.
2) For the two non-reference quantitative metrics (LPIPS, and NIQE), GFP-GAN and GPEN achieve the best or the second best individually in tasks, but none of them demonstrates a significant advantage over the other. The HiFaceGAN and PSFR-GAN do not show competitive results against these two methods and perform better than ours. 
We suspect that the observed distinct characteristics of our proposed STUNet and the compared methods can be attributed to the different nature of our method and its counterparts. All the compared methods, HiFaceGAN, PSFR-GAN, GFP-GAN and GPEN, belongs to GAN-based ones. It is well known that GAN-type methods are strong at generating content pleasing human visual perception, which will naturally gain advantages over our method in the case of non-reference quantitative metrics. Our method pays more attention to important features by computing the self-attention to introduce connections between features. Thus it outperforms other compared methods in terms of full-reference quantitative metrics.

%\fixme{this paragraph needs nearly rewrite. When you analysis, you should not keep the STUNet alone, and say which one is best, which one is second. For example, it is weird to say (face denoising) "GPEN obtains the highest scores on the SSIM, MS-SSIM, FID, and LIPIPS", especially some of the performance is actually the second best.}
%\fixme{I will revise this paragraph when the logic is smooth}

%In addition, we evaluate the performance of these methods on two task-driven metrics. As shown in Table \ref{table:128_2}, the AFLD and AFICS stand for average face landmark distance and average face ID cosine similarity. Among the four state-of-the-art methods, in terms of AFLD, PSFR-GAN achieves the best performance at the Face Deblurring and Face Artifacts Removal tasks. GPEN achieves the best performance at the Face Denoising task. HiFaceGAN achieves the best performance at the Face Super-Resolution and Blind Face Restoration tasks. In terms of AFICS, PSFR-GAN achieves the best performance at the Face Deblurring, Face Denoising and Face Artifacts Removal tasks. HiFaceGAN achieves the best performance at the Face Super-Resolution and Blind Face Restoration tasks. Based on the results, PSFR-GAN stands out as the best performer at the Face Deblurring and Face Denoising tasks. HiFaceGAN shows the best performance at the face Super Resolution and Blind Face Restoration tasks. Specially,  the proposed STUNet outperforms the existing state-of-the-art BFR methods at the  face deblurring and face denoising tasks.

In addition, we also evaluate the performance of these methods with the two task-driven metrics of AFLD and AFICS.  As shown in Table \ref{table:128_2}, regarding the AFLD metric, the best performance of different tasks is achieved by different methods, PSFR-GAN (face deblurring), STUNet (face denoising and face artifact removal), and HiFaceGAN (face super resolution and blind face restoration). The results indicate that STUNet and HiFaceGAN demonstrate the better ability in restoring LQ face images in various conditions, as they both achieve the best performance on two tasks.
Besides, PSFR-GAN achieves the best performance on the face deblurring task and achieves the second-best performance for the face artifact removal task. 
GFP-GAN and GPEN obtain only the second-best performance at one or two tasks, which are not as satisfactory as the other three methods. 
In terms of the AFICS metric, our STUNet achieves the best performance at tasks of face denoising, face artifact removal, and blind face restoration, and obtains the second best performance for the task of face super resolution. The improvement over other methods shows its advantages.
In addition, HiFaceGAN and PSFR-GAN achieve the best and second-best performance for tasks, none of them showing advance over each other.
Moreover, the GFP-GAN and GPEN fail to achieve the best or the second-best performance for any task. 
Considering that our STUNet achieves outstanding performance at three full-referenced metrics (PSNR, SSIM, and MS-SSIM), the images generated by STUNet are more similar to the original image at the pixel level and show more utility in aiding the down-stream tasks like face recognition.

To evaluate the visual quality of results by different methods, we conduct a visual comparison of these state-of-the-art methods and the proposed STUNet in Fig. \ref{fig:128_results}.  It is difficult to find the difference among images generated by PSFR-GAN, GFP-GAN, and GPEN on Noise and JPEG settings corresponding to the tasks of face denoising and face artifact removal, though the results of every evaluation metric differ from each other, as Fig. \ref{128_noise}, Fig. \ref{128_jpeg} show. Besides, on the most challenging setting Full, corresponding to the blind face restoration task, we find that the image recovered by GPEN is more satisfying to human visual perception, like Fig. \ref{128_full}. Regarding the two non-reference metrics in Table \ref{table:128_1}, for the blind face restoration task, GPEN achieves a more satisfactory performance than other methods. Therefore, we suppose that the results of the non-reference metrics are better at reflecting whether the generated image satisfies the human visual perception.
%the recovered images show the difference from the HQ image in terms of facial details and textures, like Fig. 
%\ref{128_full}. 

%\fixme{I find GPEN is the best from the qualitative perspective}

\subsection{EDFace-Celeb-150K (BFR512) Benchmark Results}

To investigate the performance of these BFR methods when given input with a resolution of $512\times512$, we carry out comparison studies among these methods and report the results in Table \ref{table:512_1}.
For the three full-referenced metrics (PSNR, SSIM, and MS-SSIM), similar to the results on the EDFace-Celeb-1M (BFR128) dataset, our STUNet is still the most competitive method among the six representative methods. STUNet achieves the best performance for face denoising, face artifact removal, and face super resolution tasks and achieves the second-best performance for the face deblurring task.
Besides, GFP-GAN also shows satisfactory performance. It achieves the best performance for the blind face restoration task and the second-best performance for the tasks of face artifact removal and face super resolution. Among DFDNet, HiFaceGAN, PSFR-GAN, and GPEN, HiFaceGAN obtains the highest scores on several metrics for tasks.
The other three methods achieve neither the best nor the second best on any task, not as well as the HiFaceGAN.
Among the two non-reference metrics (LPIPS and NIQE), regarding the LPIPS metric, GFP-GAN stands out as the best performer across all the tasks. According to the scores of the NIQE metric, DFDNet outperforms other methods for all the tasks.

%We also use the two task-driven metrics, AFLD and AFICS, to further evaluate the performance of these methods on the BFRBD512 dataset. Table \ref{table:512_2} shows the quantitative results of the two metrics. In terms of the AFLD metric, HiFaceGAN achieves the best performance at the Face Deblurring and Face Denoising tasks. PSFR-GAN achieves the best performance at the Face Artifacts Removal task. GFP-GAN achieves the best performance at the Face Super Resolution and Blind Face Restoration tasks. In terms of the AFICS metric, HiFaceGAN achieves the best performance at Face Denoising task. GFP-GAN achieves the best performance at the Face Deblurring, Face Artifacts Removal, Face Super-Resolution and Blind Face Restoration tasks. Specially, at Face Denoising, Face Artifacts Removal, and Face Super-Resolution tasks, our proposed STUNet achieves higher scores than the existing state-of-the-art methods on AFICS. 

\begin{figure*}
  \centering
    \subfigure[Visual results of different methods for the \textbf{Face Deblurring} task on the EDFace-Celeb-150K (BFR512)]{
    \label{512_blur}
    \includegraphics[width=0.95\linewidth ]{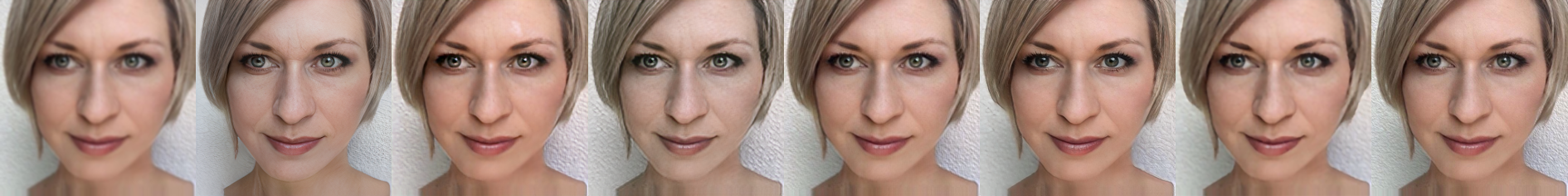}}
    \subfigure[Visual results of different methods for the \textbf{Face Denoising} task on the EDFace-Celeb-150K (BFR512)]{
    \label{512_noise}
    \includegraphics[width=0.95\linewidth ]{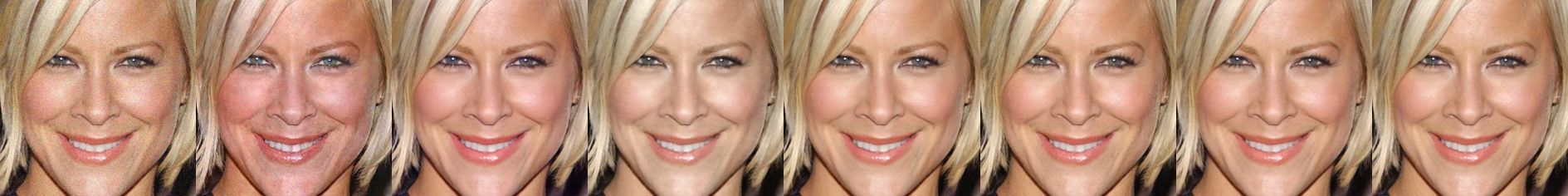}}
    \subfigure[Visual results of different methods for the \textbf{Face Artifact Removal} task on the EDFace-Celeb-150K (BFR512)]{
    \label{512_jpeg}
    \includegraphics[width=0.95\linewidth ]{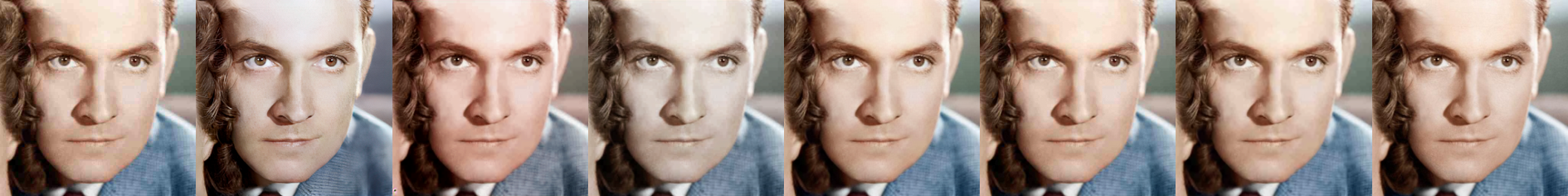}}
    \subfigure[Visual results of different methods for the \textbf{Face Super Resolution} task on the EDFace-Celeb-150K (BFR512)]{
    \label{512_lr}
    \includegraphics[width=0.95\linewidth ]{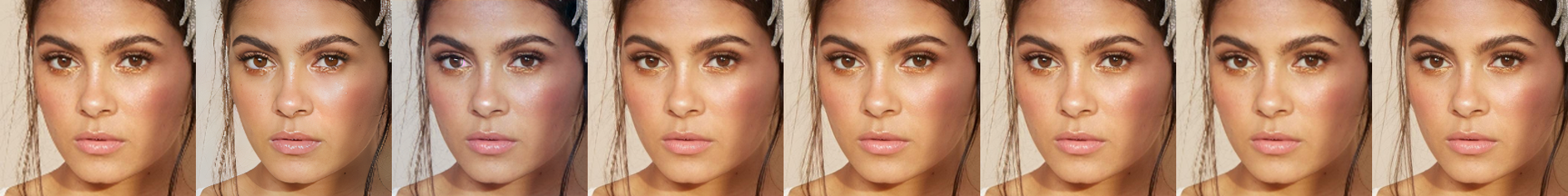}}
    \subfigure[Visual results of different methods for the \textbf{Blind Face Restoration} task on the EDFace-Celeb-150K (BFR512)]{
    \label{512_full}
   \includegraphics[width=0.95\linewidth ]{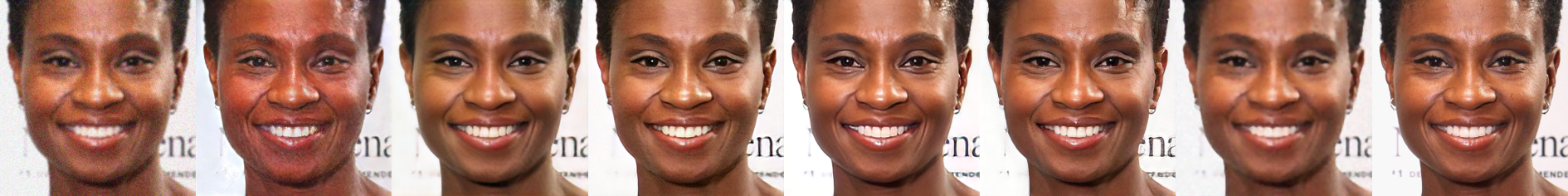}}
\caption{ \textbf{Visual results corresponding to different tasks on the EDFace-Celeb-150K (BFR512) dataset }. From left to right: LQ image, results of DFDNet\cite{li2020blind}, HiFaceGAN\cite{yang2020hifacegan}, PSFR-GAN\cite{chen2021progressive}, GFP-GAN\cite{wang2021towards}, GPEN\cite{yang2021gan}, STUNet and HQ image.}
  \label{fig:512_results} 
\end{figure*}

\begin{figure*}
  \centering
    \subfigure[Visual results of different methods with 128$\times$128 real-world input.]{
    \label{128_real}
    \includegraphics[width=0.95\linewidth ]{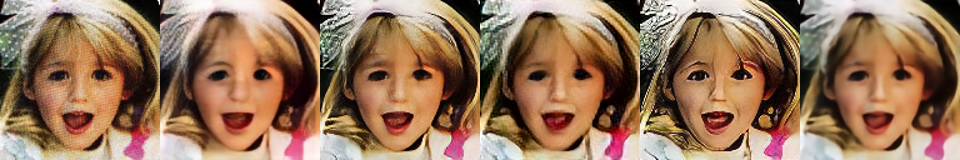}}
    \subfigure[Visual results of different methods with 512$\times$512 real-world input.]{
    \label{512_real}
    \includegraphics[width=0.95\linewidth ]{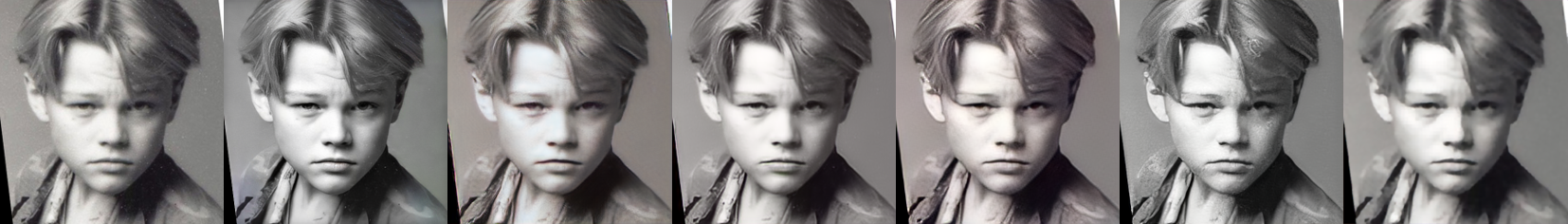}}
\caption{\textbf{Visual results of different methods on the real-world dataset}. From left to right: LQ image, results of DFDNet\cite{li2020blind} (only showed in the bottom row), HiFaceGAN\cite{yang2020hifacegan}, PSFR-GAN\cite{chen2021progressive}, GFP-GAN\cite{wang2021towards}, GPEN\cite{yang2021gan} and ours.}
  \label{fig:real} 
\end{figure*}

We also use the two task-driven metrics, AFLD and AFICS, to further evaluate the performance of these methods on the EDFace-Celeb-150K (BFR512) dataset.
Table \ref{table:512_2} shows the quantitative results of these methods.
For the AFLD metric, HiFaceGAN, PSFR-GAN, and GFP-GAN obtaining the best or the second best for different tasks perform more competitively than DFDNet, GPEN, and STUNet.
Regarding the metric of AFICS, the proposed STUNet achieves the best performance for three tasks (face denoising, face artifact removal, and face super resolution) and the second best performance for the blind face restoration task.
GFP-GAN achieves the best performance for two tasks (face deblurring and blind face restoration) and the second best performance for three tasks (face denoising, face artifact removal, and face super resolution). 
Besides, HiFaceGAN gains the second best performance for the face deblurring task. 
The other three methods do not achieve the best or the second best performance at any task. 
We suspect that the images recovered by the proposed STUNet and GFP-GAN are beneficial to improve the accuracy of face recognition.

Fig. \ref{fig:512_results} shows the visual comparison between these methods on the EDFace-Celeb-150K (BFR512) dataset. Similarly, as shown in Fig. \ref{512_noise} and Fig. \ref{512_jpeg}, for the Noise and JPEG settings, it is difficult to find the difference between the images recovered by GFP-GAN and GPEN. Though the methods have achieved strong quantitative results, there is still a gap between the recovered images and the HQ images in terms of facial details and textures, as shown in Fig. \ref{512_full}.

\subsection{Results on Real-World Dataset}
Moreover, we evaluate the performance of the representative state-of-the-art methods on a real-world dataset. The visual comparison of these methods on the real-world dataset is displayed in Fig. \ref{fig:real}. As shown in Fig. \ref{128_real}, the image recovered by GPEN has clearer facial textures, but the image recovered by PSFR-GAN is more similar to the original image. Besides, Fig .\ref{512_real} shows that the visual result of GFP-GAN is more competitive.

\section{Conclusion and Future Directions}
In this paper, we first construct two blind face restoration benchmark datasets which contain five settings corresponding to five face restoration tasks. Both the two datasets include large-scale pairs of LQ and HQ images and a fixed division of training and testing samples. Besides, we benchmark five existing blind face restoration methods on the proposed datasets and carry out a comprehensive comparison employing as many as eight quantitative metrics, including reference metrics, non-reference metrics, and task-driven metrics. Last, we propose a baseline model for blind face restoration called STUNet. The model is the first to introduce a self-attention mechanism into the blind face restoration task and achieves satisfactory performance. 

In the future, we will study more settings such as BD, and DN, where ``B" denotes blur operation, ``N" denotes the operation of adding noise, and ``D" stands for downsampling, to further explore the performance of different BFR methods. In addition, we will also explore new methods for blind face restoration.

\bibliographystyle{IEEEtran}
\bibliography{egbib}
\end{document}